\pdfoutput=1
\documentclass[10pt,twocolumn,letterpaper]{article}

\usepackage[pagenumbers]{cvpr}

\newcommand{\pn}[2]{#2}
\newcommand{\pq}[2]{#1}

\usepackage{pgfplots}
\pgfplotsset{compat=1.18}
\usepgfplotslibrary{groupplots}
\usetikzlibrary{calc,shapes.geometric}

\usepackage{colortbl}
\usepackage{enumitem}

\definecolor{cvprblue}{rgb}{0.21,0.49,0.74}
\usepackage[pagebackref,breaklinks,colorlinks,allcolors=cvprblue]{hyperref}

\hypersetup{pdftitle={Beyond Infraction Scores: Failure-Boundary Profiles for Closed-Loop Evaluation of Driving and Robot Navigation Agents},
            pdfauthor={Milan Bulat, Gino Ardelean, Ellington Kirby},
            pdfsubject={Closed-loop evaluation of driving and robot navigation agents along certified scenario axes},
            pdfkeywords={autonomous driving, closed-loop evaluation, CARLA, Isaac Sim, robot navigation, scenario generation}}
            
\def\paperID{*****} 
\def\confName{CVPR}
\def\confYear{2026}

\title{Beyond Infraction Scores: Failure-Boundary Profiles for Closed-Loop
Evaluation of Driving and Robot Navigation Agents}

\author{Milan Bulat\\
Social Chaos Lab\\
{\tt\small speedy.milan@gmail.com}
\and
Gino Ardelean\\
Social Chaos Lab\\
{\tt\small ginoardelean@gmail.com}
\and
Ellington Kirby\\
valeo.ai\\
{\tt\small ellington.kirby@valeo.com}
}

\begin{document}
\maketitle

\begin{abstract}
Closed-loop driving benchmarks in CARLA rank agents by a driving score that
multiplies route completion by an infraction penalty. This score does not
distinguish an agent that stops in front of an obstacle from one that
pushes past it. In a corridor closed by two parked cars, for example, one
agent that drives through with contact averages
\pn{fm_004_threshold_sweep.simlingo.score_split}{52.2} on the blocked widths,
while the agents that stop average between
\pn{fm_004_threshold_sweep.pdm_lite_expert.score_split}{37.8} and
\pn{fm_004_threshold_sweep.tfpp.score_split}{40.6}. We propose
failure-boundary profiles, an evaluation in which each scenario family varies
one measured factor, such as corridor width or the time gap in oncoming
traffic, and each variant is certified from measurement as one where the agent
should proceed or one where it should stop. Along this axis we estimate two
curves per agent, commitment (the probability of attempting the passage) and
competence (the probability of passing without contact), and locate where each
crosses one half, with intervals that account for the simulator's run-to-run
variance. The certified labels also give a summary metric, decision accuracy,
which credits clean passes on passable variants and stops on blocked ones. We
evaluate seven driving agents on seven CARLA families. The rule-based expert
reaches a decision accuracy of \pn{pdm_lite_expert.decision_accuracy}{0.74}
and TFv6, the highest-scoring learned agent,
\pn{lead_regnety032.decision_accuracy}{0.60}. Agents with similar driving
scores fail in different ways, and agents that reliably stop at a blocked
corridor still rarely wait for a gap in oncoming traffic. We also apply the
method, unchanged, to a warehouse robot in Isaac Sim. The navigation policy we
test stops at blocked gaps in only one of three families, and two fine-tuned
versions of it pass more gaps cleanly but no longer stop at all.
\end{abstract}
    
\section{Introduction}

Recent CARLA-based benchmarks such as Bench2Drive and
Fail2Drive~\cite{bench2drive,fail2drive} evaluate agents with the Driving 
Score (DS), which multiplies an agent's route completion by an infraction 
penalty. The route is predefined for each scenario, and the infractions cover
collisions, traffic-rule violations and blocking. The intent is that a
single number captures good driving.

These benchmarks probe a wide range of behaviors through scenario
construction: merging, obeying traffic signals, and deviating from the
usual route geometry. Bench2Drive observed that long routes entangle
different failure modes and make aggregate metrics hard to interpret, and
moved to short routes with one scenario each. We argue that even a
per-scenario driving score cannot explain why two agents differ, because it
mixes two separate questions: whether the agent attempts a maneuver, and
whether it carries it out without contact.

We therefore evaluate agents along controlled axes instead of fixed
routes. A \emph{scenario family} varies one measured factor, such as the
free width of a corridor or the time gap in oncoming traffic, and holds the
rest of the scene constant. Each variant is certified from measurement: a
corridor is passable or blocked according to its geometry, and a gap in
oncoming traffic is usable or one to wait for according to the traffic
that actually occurs. Along each axis we estimate two curves per agent.
\emph{Commitment} is the probability that the agent attempts the passage,
and \emph{competence} is the probability that it passes without contact.
The point where a curve crosses one half is a \emph{failure boundary},
which we report with an interval reflecting the simulator's run-to-run
variance. Where commitment is higher than competence, the agent keeps
attempting a passage it can no longer complete, and the width of that
region differs from agent to agent. Since every variant carries a label,
each run can also be scored as right or wrong, a clean pass where passage
is possible and a stop where it is not. We call the resulting summary
\emph{decision accuracy}. We apply this to seven driving agents on seven 
families, producing curves that show each agent's commitment-competence 
gap, and we analyze the failures behind it. 

The same curves and the same decision accuracy carry over to robot
navigation without modification. We test a world-model-based navigation 
policy in simulation, producing similar failure-boundary curves, and 
demonstrate the effects of a simple fine-tuning strategy.

Our contributions are:
\begin{enumerate}[(i),nosep]
  \item a pipeline that turns natural-language requests into scenario
        families whose variants differ in one measured factor and are
        certified passable or blocked, with the spawned geometry checked in
        the simulator;
  \item failure-boundary profiles, i.e.\ commitment and competence curves
        along a certified axis, with boundaries and intervals that account
        for the simulator's run-to-run nondeterminism;
  \item decision accuracy, a summary metric that credits a clean pass on
        variants certified passable and a stop on variants certified
        blocked;
  \item an evaluation of seven driving agents on seven CARLA families and
        of a warehouse robot on three Isaac Sim families. Agents with
        similar driving scores fail in different ways, within a family the
        driving score can rank an agent that never stops above agents that
        stop correctly, and learned agents rarely wait for a gap in
        oncoming traffic, including those that stop at a blocked corridor.
  \item the ten scenario families, the measurement chain and the records of
        all seventeen campaigns in a
        \href{https://github.com/speedym/failure-boundary-profiles}{public repository},
        and the simulator evidence the scores were based on in a
        \href{https://doi.org/10.57967/hf/11195}{dataset on the Hugging Face Hub}.   
\end{enumerate}

\section{Related Work}
We review closed-loop evaluation of driving agents, the evaluation of robot
navigation, and scenario generation.

\paragraph{Closed-loop evaluation of generalization.}
Most closed-loop evaluation of end-to-end driving runs in the CARLA
simulator~\cite{carla}. The CARLA Leaderboard~\cite{carla_leaderboard}
scores agents on long predefined routes with the driving score, the product
of route completion and an infraction penalty, and Leaderboard~2.0 extends
the routes with a large set of interactive scenarios.
Bench2Drive~\cite{bench2drive} argues that long routes entangle failure
modes and replaces them with short routes containing one scenario each,
reporting driving score and success rate per ability.
Fail2Drive~\cite{fail2drive}, on which this work builds most directly,
pairs in-distribution routes with shifted counterparts in which only the
targeted factor changes
(\pq{200}{Fail2Drive (Gerstenecker et al., 2026), its route count} routes in
all), and reports an average success-rate drop of
\pq{22.8}{Fail2Drive (Gerstenecker et al., 2026), its reported average drop}\%
for state-of-the-art agents under these shifts. We reuse its framework,
several of its scenario classes and its extended expert, and evaluate an
overlapping set of agents.

Our work differs from Fail2Drive in five respects. Fail2Drive measures
whether performance drops between two conditions, whereas we sweep a
measured factor over many certified variants and locate, with an interval,
where an agent's behavior changes. Fail2Drive reports driving score,
success rate and their harmonic mean; we show that the driving score can
rank an agent that forces its way past a blockage above one that stops,
and we add a decision accuracy computed against certified labels.
Fail2Drive uses the expert's performance as its passability check, while we
certify each variant from its own measurement and check it offline and in the
simulator. The difference matters in practice: on our machine-intrusion
family the expert hits the obstacle in every run although every variant is
passable, so an expert-based check would have labeled the family blocked.
We also include impassable variants on purpose, so that an agent cannot
score well by committing everywhere. Finally, the oncoming-flow families
add time as an axis, by varying the gap available to cross.
PlanT~\cite{plant,plant2} studies generalization failures with an
object-centric representation and shows a lack of robustness to simple
object perturbations. Simulators based on neural
rendering~\cite{zhou2024hugsim,alpasim_2025} bring realistic sensor input
to closed-loop evaluation and are a natural next step, but their cost
currently makes the repeated runs that boundary profiling needs too
expensive, and certifying their geometry is still an open problem.

\paragraph{Robot navigation evaluation.}
Mobile robot navigation has the same aggregation problem.
BARN~\cite{Perille_2020} benchmarks ground navigation through cluttered
environments of graded difficulty, which is the closest existing analog to
a swept axis. Metrics based on reaching the goal have the same weakness as
the driving score: when the goal cannot be reached through a gap, a
correct stop in front of it and a failure look the same. We evaluate
X-Mobility~\cite{xmobility}, an end-to-end navigation policy trained with
world modeling, on a warehouse robot in Isaac Sim~\cite{isaacsim}. Each gap
is certified passable or blocked and a detour through a neighboring aisle
is always available, so stopping, waiting and going around can all be
scored.

\paragraph{Scenario generation.}
Parameterized scenarios are standard in CARLA through
ScenarioRunner~\cite{scenariorunner} and OpenSCENARIO, and probabilistic
languages such as Scenic~\cite{scenic} describe distributions over scenes.
A large body of work generates safety-critical scenarios adversarially,
perturbing traffic participants until the agent fails~\cite{king,advsim},
and recent work writes scenarios from natural language with large language
models~\cite{chatscene,lctgen}. Closest to our goal, performance-boundary
search in autonomous-vehicle testing looks for the region of a parameter
space where a system's behavior changes~\cite{mullins2018}. Failure search
finds individual failures; we measure where behavior changes along a
certified axis, with its uncertainty, and compare that location across
agents. Our pipeline also gives different guarantees. A constrained
planner turns a natural-language request into a declarative plan, a
deterministic solver builds each variant, and each variant is
measured and checked, offline and in the simulator. 
Every variant therefore carries a certified label. A second check on the
recorded runs catches problems that the offline measurement cannot, such
as oncoming cars that collide with each other (\autoref{sec:pipeline}).

\section{Method}
In \autoref{sec:case_study} we first motivate the introduction of failure-boundary profiling with an illustrative case study. In \autoref{sec:metrics} we describe newly developed metrics to explore failure-boundary profiles. In \autoref{sec:profiles} we introduce the new set of scenario families, one failure-boundary profile each, created by the Social Chaos Lab (SCL) pipeline. Finally, in \autoref{sec:pipeline} we illustrate the steps of the SCL scenario generation pipeline.
\subsection{A Case Study in Failure Profiling}
\label{sec:case_study}
\begin{figure}[t]
    \centering
    \includegraphics[width=0.9\linewidth]{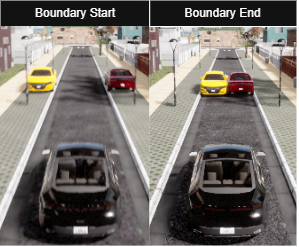}
    \caption{\textbf{The two ends of the Charger gauntlet.} Left: the widest
    variant, with two cars parked on the shoulders and a
    \pn{fm_004_threshold_sweep.axis_range}{3.64}\,m free corridor between
    them. Right: the narrowest variant, where the two cars close the corridor
    completely.}
    \label{fig:case_study}
\end{figure}

We consider a constructed scenario in which route completion and infraction score become opposite objectives. 

In this scenario the ego vehicle must pass between two cars parked on the two shoulders of the road. Neither car impinges upon the drivable surface, and thus the vehicle should drive straight through. 

We then systematically reduce the spacing between these two vehicles, thereby reducing the drivable corridor. \autoref{fig:case_study} shows the two ends of this scenario. By repeating evaluations of the agent at each reduction in corridor width, we obtain a curve over the variations in agent performance. 

\begin{figure}[t]
\centering
\includegraphics[width=\linewidth]{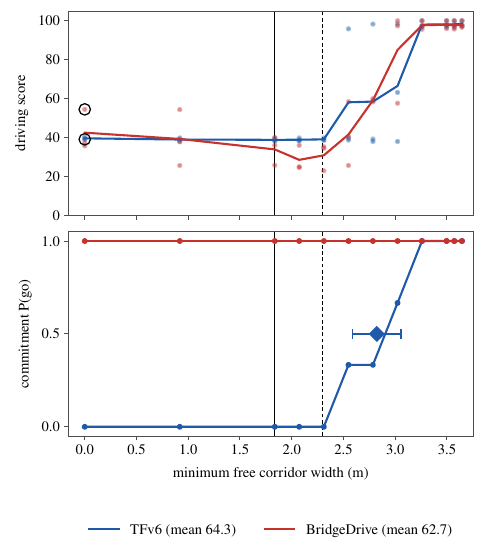}
\caption{\textbf{Driving score and commitment on the Charger gauntlet.} TFv6
and BridgeDrive, three traffic seeds each, against the minimum free corridor
width. Top: driving score of each run (dots) and the mean per width (lines);
the circled pair is the same width and seed, the fully closed corridor at seed
2, where TFv6 stops
(\pn{fm_004_threshold_sweep.lead_regnety032.closed_corridor_scores}{39.1}) and
BridgeDrive continues past the blockage with contact
(\pn{fm_004_threshold_sweep.bridgedrive.closed_corridor_scores}{54.5}).
Bottom: commitment per width; the diamond marks TFv6's commitment boundary
with its bootstrap interval. Solid line: body width,
\pn{fm_004_threshold_sweep.body_limit_m}{1.84}\,m; dashed line: policy floor,
\pn{fm_004_threshold_sweep.policy_floor_m}{2.30}\,m.}
\label{fig:case_study_graph}
\end{figure}

\autoref{fig:case_study_graph} shows two baseline agents (TFv6~\cite{lead} and BridgeDrive~\cite{bridgedrive}) varied over corridor width. We measure two quantities across repetitions: the CARLA Driving Score, and a new measure detailed further in \autoref{sec:metrics}, the probability that the agent attempts to continue along the route.

\autoref{fig:case_study_graph} illustrates the tension between route completion and infraction score: the agent can win a higher driving score by trading infractions for progress. Looking at commitment, we can see a clear boundary at which point the TFv6~\cite{lead} agent refuses to attempt the gap. The BridgeDrive~\cite{bridgedrive} agent meanwhile attempts the gap no matter the width.

\begin{table}[t]
\centering
\footnotesize
\setlength{\tabcolsep}{3.5pt}
\begin{tabular}{@{}lrrrrrr@{}}
\toprule
 & \multicolumn{3}{c}{Mean DS} & \multicolumn{3}{c}{Blocked widths} \\
\cmidrule(lr){2-4}\cmidrule(l){5-7}
Agent & All & Pass. & Block. & $n$ & Stop & Contact \\
\midrule
PlanT 2.0$^{\ast}$ & 72.5 & 86.7 & 44.0 & 12 & 3  & 6 \\
HiP-AD             & 71.5 & 87.7 & 39.2 & 12 & 11 & 1 \\
SimLingo           & 67.8 & 75.5 & 52.2 & 12 & 0  & 12 \\
TF++               & 65.6 & 78.1 & 40.6 & 12 & 12 & 0 \\
TFv6               & 64.3 & 76.9 & 39.1 & 12 & 12 & 0 \\
BridgeDrive        & 62.7 & 76.0 & 36.1 & 12 & 0  & 12 \\
Expert             & 59.2 & 69.9 & 37.8 & 12 & 12 & 0 \\
\bottomrule
\end{tabular}

\caption{\textbf{Driving score split by certified passability.} Mean
driving score on the Charger gauntlet (three traffic seeds) over all runs, over the eight
passable widths and over the four blocked widths
(\pn{fm_004_threshold_sweep.blocked_axis_range}{0.00 to 2.07}\,m), with the
outcomes of the runs on the blocked widths. Stop: the agent stops before the
blockage. Contact: any contact with the blockage, whether the agent continued
past it or not. Rows are ordered by the mean over all runs.
$^{\ast}$Checkpoint 1; its other
\pn{fm_004_threshold_sweep.plant2.blocked.anomaly_blockage_completed}{3}
blocked runs complete the
\pn{fm_004_threshold_sweep.blocked_axis_range}{2.07}\,m width without
contact.}
\label{tab:case-study-split}
\end{table}
This curve represents a failure-boundary profile: a measure of an agent's behavior over a specific, controlled change. This type of evaluation in simulation allows fine-grained analysis of an agent's generalization capacity over specific criteria. 

Creating this type of scenario family would typically be a tedious process of iterative work in the simulator. In \autoref{sec:pipeline} we detail the automated pipeline built to generate these scenario families.

\subsection{Measuring Failure-Boundary Profiles}
\label{sec:metrics}
This section defines the two curves of a failure-boundary profile, the boundary on each, and a summary metric, decision accuracy.

\newcommand{\Pgo}{P_{\mathrm{go}}}
\newcommand{\Pclean}{P_{\mathrm{clean}}}

Let a scenario family $\mathcal{F}$ schedule one certified axis
$x \in \mathcal{X} \subset \mathbb{R}$ (e.g.\ minimum free corridor width in
meters, or crossing time gap in seconds), realized at variants
$x_1 < \dots < x_V$. For agent $a$, each closed-loop run $r$ at variant $x_v$
under traffic seed $s$ yields an outcome $o_r \in \mathcal{O}$ from the
family-aware outcome taxonomy, the set of run outcomes the scorer
distinguishes for that family (\autoref{sec:pipeline}). Two binary events
are read from $o_r$:
\begin{align}
  G_r &= \mathbb{1}\{\text{agent } a \text{ attempts the passage in run } r\},
  &&\text{(a \emph{go})}\\
  C_r &= \mathbb{1}\{o_r = \textsc{clean pass}\}.
\end{align}

\textbf{Commitment}
The commitment curve of agent $a$ on family $\mathcal{F}$ is
\begin{equation}
  \Pgo^{a}(x) \;=\; \Pr\big[\,G = 1 \,\big|\, x,\; a,\; \mathcal{F}\,\big],
\end{equation}
the probability that the agent commits to the passage at axis value $x$.

\textbf{Competence}
The competence curve of agent $a$ on family $\mathcal{F}$ is
\begin{equation}
  \Pclean^{a}(x) \;=\; \Pr\big[\,C = 1 \,\big|\, x,\; a,\; \mathcal{F}\,\big],
\end{equation}
the probability that the agent passes cleanly at axis value $x$.

The two curves are estimated separately, each over the runs that can be
scored for it. A late commitment, where the agent passes only after the
scenario's time limit has expired, counts as a go but is left out of
competence, so the estimated curves need not satisfy $\Pclean \le \Pgo$ at
every $x$.

\textbf{Boundary}
For a curve $P \in \{\Pgo, \Pclean\}$ with the family's orientation
$\sigma \in \{+1,-1\}$ (passage easier as $\sigma x$ grows), the boundary is
the $p = \tfrac12$ crossing of the fitted curve, for $\sigma = +1$
\begin{equation}
  x^{\star} = \inf\{\,x \in \mathcal{X} : \hat{P}(x) \ge \tfrac12\,\},
\end{equation}
and for $\sigma = -1$ the supremum of the same set, estimated by a logistic fit
$\hat{P}(x) = \big(1 + e^{-\beta\sigma(x - x^{\star})}\big)^{-1}$.
Four cases arise. When at least one variant has mixed outcomes, the fit is
constrained and the boundary is its midpoint. When every failing variant
lies on one side of every passing one, there is nothing to fit, and the
boundary is the midpoint of the two adjacent variants between which the
outcome flips, reported as a separation bracket
$[x_{\mathrm{fail}}, x_{\mathrm{pass}}]$. When the fit does not converge,
as with a single mixed variant between the two sides, the bracket of the
last failing and the first passing variant takes its place. When the
observed rate never crosses one half in the tested range, no boundary is
reported, and the panel title names the case (``flat'', ``not located in
range'' or ``below tested range'').
We treat each run as an independent Bernoulli draw and pool the traffic
seeds within a variant as a stratum, since a fixed seed does not reproduce a
run (\autoref{sec:variance}). A fitted boundary carries a bootstrap interval
that resamples the runs within each variant holding more than one; a
bracket is its own interval; a flat curve carries a Wilson interval on its
level.

The \emph{commitment-competence gap},
$\Delta^{a}(x) = \Pgo^{a}(x) - \Pclean^{a}(x)$, covers the axis values
where the agent still commits but no longer passes cleanly; the profile 
figures draw it in orange. Where this region lies, and how wide it is, 
differs considerably between agents.

\textbf{Decision Accuracy}
Commitment and competence describe what an agent does. Decision accuracy
compares this with what it should have done, using the label that every
variant carries. On the static-obstacle and warehouse families the pipeline 
certifies each variant from its measured geometry as passable or blocked 
against a declared floor: for the vehicle the policy floor, its body width 
plus a mirror allowance and a clearance margin on each side; for the robot 
its width measured at each height (\autoref{sec:pipeline}). On the oncoming-flow 
families a gap certificate labels each variant and seed from the realized 
traffic as usable, conditional or wait (\autoref{sec:pipeline}).
The correct outcome is a clean pass on a passable or usable variant and a
stop on a blocked or wait one. A go that ends in contact does not count as
correct, since an agent that cannot complete a passage should not attempt
it. For agent $a$ on family
$\mathcal{F}$, with $C = 1$ for a passage without contact,
\begin{equation}
  \begin{split}
  \mathrm{DA}^{a}_{\mathcal{F}} = \tfrac12\Bigl(
    &\Pr\big[\,G = 1, C = 1 \,\big|\, \text{passable}\,\big]\\
    &+ \Pr\big[\,G = 0 \,\big|\, \text{blocked}\,\big]\Bigr),
  \end{split}
\end{equation}
the balanced accuracy over the two classes. A late commitment counts as a
miss on the passable side, and anomalous runs, those that complete a variant
certified blocked without contact, are excluded. An agent that
always stops scores $\tfrac12$; an agent that always goes scores at most
$\tfrac12$, and less for every passable variant it does not pass cleanly. 
The per-agent summary averages $\mathrm{DA}^{a}_{\mathcal{F}}$ with equal 
weight over the families that contain a stop side; families with only 
passable variants do not enter it. Its \pn{bootstrap.level_percent}{95}\% 
interval is a bootstrap that resamples runs within each family and side
(\pn{bootstrap.replicates}{10,000} replicates). Conditional flow variants,
where only a rolling crossing fits the gap, are reported separately and enter
neither side.

We report decision accuracy alongside the driving score.
The driving score mixes route completion with infraction penalties, so on a
blocked variant continuing past the obstacle with contact can score higher
than stopping in front of it (\autoref{tab:case-study-split}). The two
labels differ in kind: a blocked width is fixed by geometry against the
policy floor, whereas a wait window is a timing rule the
certificate states, a kinematic floor plus a declared margin. Both are
conventions with declared constants; \autoref{sec:results-overview} reports
how the ranking responds to each.
\subsection{Scenario Families}
\label{sec:profiles}
\begin{figure*}[t]
\centering
\includegraphics[width=\textwidth]{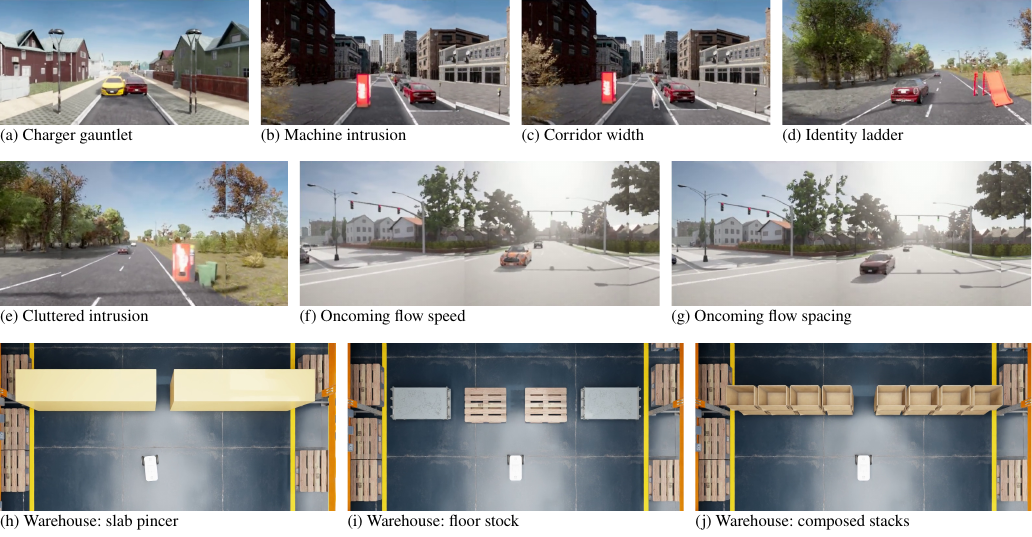}
\caption{\textbf{The ten scenario families.} One picture per family, cut
from the recordings of the campaigns the paper reports: (a) to (g) the front
cameras of TFv6 in the CARLA families, (h) to (j) the top-down view of the
warehouse families. (a) The Charger gauntlet at its closed end, the two parked
cars meeting across the lane. (b) The vending machine at the lane edge, the
shallowest intrusion of the sweep. (c) The narrowest corridor: two street
barriers in a row on the right, opposite the vending machine on the left
curb. (d) The identity ladder with the slide in
place. (e) The cluttered intrusion, the machine and two bins at the lane
edge. (f) The slowest oncoming flow and (g) a mid-range spacing, the car waiting
at the junction to turn left across the oncoming lane. (h) The slab pincer
at a gap certified blocked, (i) the floor stock and (j) the composed stacks
at their narrowest passable gaps, the robot approaching the row.}
\label{fig:families}
\end{figure*}

We use the SCL pipeline (\autoref{sec:pipeline}) to author ten scenario families, each measuring a different failure-boundary profile: seven driving families in CARLA and three warehouse families in Isaac Sim. \autoref{fig:families} shows one picture of each.
\begin{enumerate}
    \item \textbf{Charger Gauntlet:} Two parked Dodge Chargers narrow the
    corridor across the passability threshold, with the narrowest variants
    certified blocked. Axis: minimum free corridor width,
    \pn{fm_004_threshold_sweep.variants}{12} variants,
    \pn{fm_004_threshold_sweep.repetitions_per_variant}{3} seeds.
    \item \textbf{Machine Intrusion:} A vending machine intrudes on the lane.
    Axis: intrusion depth in meters, \pn{fm_001_intrusion_sweep.variants}{12}
    variants, \pn{fm_001_intrusion_sweep.repetitions_per_variant}{3} traffic
    seeds.
    \item \textbf{Corridor Width:}  A vending machine stands on the left curb
    and two street barriers, moved in from the right, pinch the corridor
    between them.
    Axis: minimum free corridor width in meters,
    \pn{fm_002_corridor_pincer.variants}{10} variants,
    \pn{fm_002_corridor_pincer.repetitions_per_variant}{3} seeds.
    \item \textbf{Identity Ladder:} Fourteen roadside objects at one certified
    intrusion, \pn{fm_003_asset_swap.max_route_intrusion_m}{0.5}\,m, with about
    \pn{fm_003_asset_swap.min_free_corridor_m}{8}\,m free corridor. Axis:
    object identity with geometry fixed, \pn{fm_003_asset_swap.variants}{14}
    variants, \pn{fm_003_asset_swap.repetitions_per_variant}{3} seeds.
    \item \textbf{Cluttered Intrusion:} Vending machine plus two waste bins
    arranged irregularly. Axis: maximum route intrusion,
    \pn{fm_005_hero.variants}{12} variants,
    \pn{fm_005_hero.repetitions_per_variant}{3} seeds.
    \item \textbf{Oncoming Flow Speed:} Signalized left turn across a
    single-lane oncoming flow at the Town03 junction, flow speed swept with
    spacing fixed. Axis: crossing time gap,
    \pn{fm_011_flow_speed_sweep.variants}{11} variants,
    \pn{fm_011_flow_speed_sweep.repetitions_per_variant}{3} seeds.
    \item \textbf{Oncoming Flow Spacing:} Signalized left turn, spacing
    between oncoming cars swept with speed fixed. Axis: crossing time gap,
    \pn{fm_012_source_dist_interval_sweep.variants}{11} variants,
    \pn{fm_012_source_dist_interval_sweep.repetitions_per_variant}{3} seeds.
    \item \textbf{Warehouse: Slab Pincer:} Two abstract heavy slabs narrow a
    warehouse aisle to a single central gap. Axis: minimum free gap,
    \pn{fm_isaac_001.variants}{12} variants from
    \pn{fm_isaac_001.axis_range}{2.5}\,m to a closed aisle,
    \pn{fm_isaac_001.repetitions_per_variant}{5} policy samples.
    \item \textbf{Warehouse: Floor Stock:} Pushcarts and pallets already in
    the warehouse are repositioned to narrow the aisle. Axis: minimum free
    gap, \pn{fm_isaac_003.variants}{12} variants,
    \pn{fm_isaac_003.repetitions_per_variant}{5} policy samples.
    \item \textbf{Warehouse: Composed Stacks:} Cartons stacked as a picker
    would leave them narrow the aisle. Axis: minimum free gap,
    \pn{fm_isaac_004.variants}{10} variants,
    \pn{fm_isaac_004.repetitions_per_variant}{5} policy samples.
\end{enumerate}
In every warehouse family the neighboring aisle stays open, so stopping or
detouring is always a fair choice, and the narrowest gaps are certified
blocked: four on the slab pincer and the floor stock, three on the composed
stacks, where \pn{fm_isaac_004.candidates.rejected.gap_in_dead_band}{two}
further widths lie too close to the robot's width to be labeled passable or
blocked at the measurement's resolution and are set aside
(\autoref{sec:pipeline}).
\subsection{The Scenario Generation Pipeline}
\label{sec:pipeline}
\begin{figure*}[t]
    \centering
    \includegraphics[width=\linewidth]{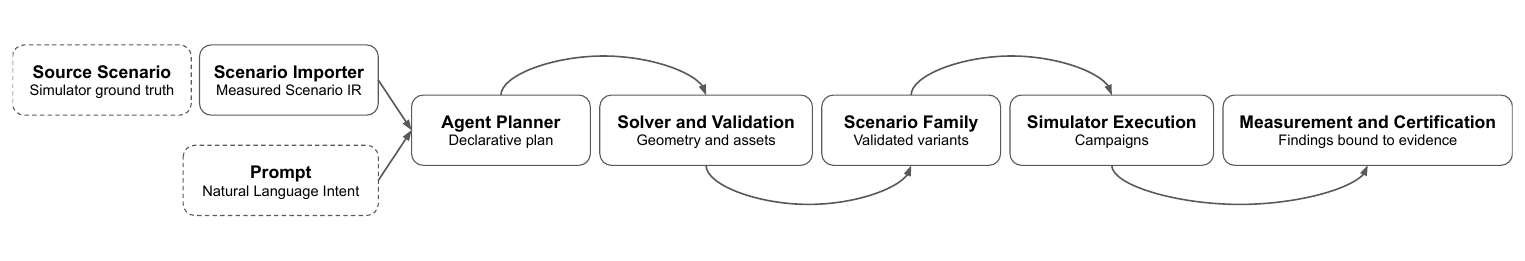}
    \caption{\textbf{The Social Chaos Lab scenario-generation pipeline.} A
    source scenario and a natural-language intent enter the pipeline; an
    agent planner turns the intent into a declarative plan, a deterministic
    solver realizes and validates each variant, and the accepted variants
    form a scenario family whose placed objects are countersigned in the
    simulator before campaigns run; campaigns are then run, scored and
    certified.}
    \label{fig:pipeline}
\end{figure*}

\autoref{fig:pipeline} shows the pipeline. It takes a source scenario and a
natural-language intent and produces a scenario family whose variants differ
in exactly one measured factor, then runs, scores and certifies campaigns on
it. The stages are as follows.

\textbf{Scenario importer.} A source scenario (for example a Fail2Drive scenario on
a Leaderboard~2.0 route, or a manually created scenario) is imported into a measured scenario: the route,
the placed actors and the drivable envelope, with curb lines measured from
the map rather than taken from the OpenDRIVE bounds.

\textbf{Agent planner.} The prompt states the intent in plain language: which
factor to sweep, over what range, and what to hold constant
(\autoref{sec:profiles}). The planner turns
it into a declarative plan under a capability profile that fixes the ego's
policy constants, such as the mirror allowance and clearance margin.

\textbf{Solver and validation.} A deterministic solver realizes each
variant from the plan, locates the factor's feasible domain by scan and
bisection over the feasibility predicate, and densifies the schedule where a
transition is expected. Each variant is measured (minimum free corridor
width, maximum route intrusion, required lateral deviation, closest
approach; realization tolerance $\pm5$\,mm), assigned passable or blocked
from that geometry, and checked by validators for collision-free placement,
stable support, connected ego passage and family membership. A candidate
that fails is rejected.

\textbf{Scenario family.} The accepted variants form the family: a manifest
with the measured axis value per variant and one compiled route file each.
Before any campaign each placed object is countersigned in the simulator: spawned
headless, settled and read back, with at most
\pq{10}{test_family_closure.py CASES, planar bar; vertical up to 120 mm}\,mm
of planar drift between authored and spawned pose. The countersignature helps 
catch misplaced objects: of \pn{fm_003_asset_swap.candidates}{23} candidate 
objects for the identity ladder, seven pass every offline check and then tip, 
roll, sink or fall through the road under run-time physics, and are rejected; 
two more fail the offline checks, leaving the fourteen objects of the family.

\textbf{Flow certification.} The oncoming-flow variants have no geometric
passability, and traffic cannot be countersigned the way a placed object
can. They are certified in four layers instead, three of them after the
runs, from the recordings.
(i)~\emph{At generation}, the manifest declares each variant's nominal
factor, the flow speed or the spawn interval, and the crossing gap derived
from it; the validators check family membership and that the route
compiles. The nominal gap only identifies the variant and is never used as
a measurement.
(ii)~\emph{The conflict lane.} A scoring file stores the polygon of the
junction's conflicting driving lane, built from the map. A separate check
runs it against CARLA's offline map API: the connector lane exists exactly
once in the declared junction, every recorded ego spawn reaches it, every
flow car spawns on an incoming lane the map declares, and every declared
cell holds exactly its variants' recordings.
(iii)~\emph{The gap certificate.} For each variant and seed, the
certificate measures what the traffic did: each car's arrival at and
departure from the ego's crossing region, the part of the conflict lane its
body sweeps on a reference clean crossing. From these it derives the local
headways and the occupancy of the crossing. Every run is read only until
the ego enters the oncoming lane. The usable window is the median headway 
minus the median occupancy. It is compared with two floors measured from 
the vehicle's recordings in the first four campaigns of each family, each 
plus a declared margin of
\pn{flow_families.certificate.margin_s}{0.5}\,s. The first is the time to
cross from rest, covering the standstill-to-exit distance at the vehicle's
peak acceleration of
\pn{flow_families.certificate.peak_accel_mps2}{7.04}\,m/s$^2$:
\pn{fm_011_flow_speed_sweep.certificate.from_rest_s}{2.02}\,s on the speed
sweep and
\pn{fm_012_source_dist_interval_sweep.certificate.from_rest_s}{1.81}\,s on the
spacing sweep. The second is the time of a rolling crossing at the recorded
entry speed: \pn{flow_families.certificate.rolling_reference_s}{1.00}\,s on
both sweeps. A window that fits a crossing from rest is usable, one that fits
only a rolling crossing is conditional, and one that fits neither is a wait
window; a variant whose seeds disagree is mixed and is labeled per seed, and a
seed in which no car completed its passage through the crossing before the
ego entered has no measured window and enters neither side. The
comparison always uses the realized window rather than the nominal one,
because short gaps stretch on congested seeds. Since the label depends only on
the traffic, a wait and a collision on the same variant and seed are judged
against the same window. (iv)~\emph{The defect register.} A variant whose
realized traffic differs from the declared traffic is removed from the bands
by a declared entry that records the evidence. This happens once, for the
shortest spacing variant: its cars, spawned
\pn{fm_012_source_dist_interval_sweep.defect_v01.spawn_headway_s}{0.45}\,s
apart, collide with each other and reset before the ego arrives in every run,
and the variant is excluded from every score  (grey in
\autoref{fig:profiles-spacing}).

Scoring uses all four layers. A flow collision is a contact with a
scenario actor, and a stall is a timeout at the junction. Commitment and
competence are fitted on the measured gap. Decision accuracy credits a
clean crossing of a usable window and a wait at a wait window
(\autoref{sec:metrics}), while conditional windows are scored from the
recorded entry state and reported separately. Every scoring document
references the conflict lane, the certificate and the defect register by
hash. The certificate has three limits. A wait window is a rule that it
states, which is weaker than the physical impossibility of a blocked
width. A platoon whose cars leave no gap between them is a wait window by
construction rather than by measurement. And the flow can only be certified 
after the campaign has run.

\textbf{Warehouse certification.} The warehouse families pass through the
same stages with a different backend. Each variant is written as an Isaac
Sim stage, and its free gap is measured on an occupancy grid at
\pn{warehouse.grid_res_m}{1\,cm} resolution built from the stage. The label
is then decided per height slab: the grid holds one layer per centimeter of
height with the collision geometry inside that slab, each layer is eroded by
the robot's width measured at that height (in the simulator, by closing two
walls on the robot slab by slab until they touch; at most
\pn{warehouse.robot_contact_width_m}{0.494}\,m, at the wheels), and a
connectivity test certifies whether the robot's center has a
path through the gap that is free in every slab at once, and whether it can
detour through the neighboring aisle. A gap too close to the robot's width for
the measurement to decide at its resolution is refused and reported rather
than labeled, and its runs are set aside from every curve. Stacked stock is
placed as dynamic bodies that settle under
physics; the settled poses are read back and countersigned before any run,
and the parts stand as static colliders at those poses during a run. The
simulation is audited
bit-exact, so repeated runs sample the policy, not the simulator. All
warehouse results come from one simulator vintage (Isaac Sim 5.1 on one GPU
type), and results from different vintages are never pooled.

\textbf{Simulator execution.} A campaign declares its grid (variants, agents,
repetitions) and runs it with recordings on. Because closed-loop outcomes are not deterministic at a fixed
traffic seed (\autoref{sec:variance}), each run is treated as one independent 
draw and a variant's seeds are pooled.

\textbf{Measurement and certification.} Runs are classified by the
family-aware outcome taxonomy, collisions are attributed to obstacle or
traffic from the recorder, trajectory metrics are extracted, and the
boundary estimator of \autoref{sec:metrics} is run per agent and family.
\section{Experiments}
\autoref{sec:env} describes the simulators, agents and protocol, and
\autoref{sec:variance} the run-to-run variance the protocol accounts for.
\autoref{sec:results} reports the results family by family, and
\autoref{sec:results-training} re-measures one agent across training.
\subsection{Evaluation Environment and Baselines}
\label{sec:env}
Driving agents are evaluated in the CARLA simulator~\cite{carla}, the standard open-source driving simulator, on scenarios that build on the Bench2Drive and Fail2Drive families. The warehouse robot is evaluated in NVIDIA Isaac Sim~\cite{isaacsim} on the three warehouse families. Each agent is evaluated on every family of its domain, with the two exceptions noted below.
\paragraph{Baselines.}
\begin{itemize}
  \item \textbf{TFv6}~\cite{lead}: the sensor agent of the LEAD paper, a transformer-based planner in the TransFuser lineage,
  with one wide camera, LiDAR and radar. We evaluate its flagship
  configuration with a RegNetY-032 image backbone, the configuration whose
  published scores the Fail2Drive comparison reports.
  \item \textbf{BridgeDrive}~\cite{bridgedrive}: an anchor-guided diffusion
  bridge planner. TFv6 and BridgeDrive are the two top
  agents on Fail2Drive.
  \item \textbf{TF++}~\cite{tfpp}: the TransFuser++ sensor agent shipped
  with the Fail2Drive framework.
  \item \textbf{PlanT~2.0}~\cite{plant2}: a privileged planner operating on
  object-level inputs rather than sensors.
  \item \textbf{SimLingo}~\cite{simlingo}: a vision-language driving agent
  on an InternVL2-1B backbone.
  \item \textbf{HiP-AD}~\cite{hipad}: a six-camera end-to-end agent which uses a multi-granularity planning query representation and deformable attention.
  \item \textbf{Expert (PDM-Lite)}~\cite{pdmlite}: the privileged rule-based autopilot
  shipped with the Fail2Drive framework. We use it as a reference rather than as a 
  competing agent, because its obstacle avoidance is a scripted route shift around the
  registered obstacle. It serves as a feasibility witness and a
  conservative reference line. It does not register the corridor-width family's barriers
  and is absent from that family.
\end{itemize}
Every agent is evaluated at three traffic seeds per variant (seeds 0 to 2);
where a campaign ran more seeds, only the first three are used. PlanT~2.0
did not run the identity ladder.

\paragraph{Warehouse agents.} The robot is a wheeled warehouse platform
driven by \textbf{X-Mobility}~\cite{xmobility}, an end-to-end navigation
policy, run as its released checkpoint and as two fine-tunes of it
(\autoref{sec:results-training}). A scripted \textbf{expert} that reads the stage serves
as the warehouse reference: it takes every passable gap and detours through
the neighboring aisle at every blocked one. Isaac Sim is audited bit-exact
on our setup, so repeats sample the policy rather than the simulator; each
gap is run with five policy samples.
\subsection{Variance in CARLA}
\label{sec:variance}
CARLA outcomes are not repeatable, even at a fixed traffic seed, a known
property of game-engine simulators~\cite{nondeterminism2021}.  Running the
same variant and seed five times, twice from a byte-identical command,
produces two different outcomes, with driving scores of
\pn{bifurcation.branch_53_7.driving_score}{53.7} and
\pn{bifurcation.branch_32_2.driving_score}{32.2}; the
runs match exactly until a collision, where the physics engine diverges. A
single run is therefore only one sample of the outcome. We treat every run as
an independent draw, run each variant under several seeds and pool them within
the variant, and report each boundary with a bootstrap interval that
resamples the runs within each variant, so that the simulator's noise widens
the interval instead of moving the estimate. When the behavior is stable the intervals are tight; the
expert's gauntlet boundary, for instance, is identical at all three seeds.
\subsection{Results}
\label{sec:results}
\begin{table}[t]
\centering
\footnotesize
\setlength{\tabcolsep}{5pt}
\begin{tabular}{@{}lrl@{}}
\toprule
Agent & Avg.\ DS & Avg.\ DA [95\% CI] \\
\midrule
\multicolumn{3}{@{}l}{\emph{Driving (CARLA)}} \\
Expert (reference) & 79.8          & 0.74 [0.67, 0.81] \\
TFv6               & \textbf{75.5} & \textbf{0.60} [0.53, 0.68] \\
PlanT 2.0 (ckpt 1) & 72.6          & 0.56 [0.50, 0.61] \\
PlanT 2.0 (ckpt 3) & 71.4          & 0.50 [0.50, 0.50] \\
TF++               & 59.5          & 0.49 [0.41, 0.57] \\
PlanT 2.0 (ckpt 2) & 70.4          & 0.47 [0.42, 0.50] \\
HiP-AD             & 70.4          & 0.44 [0.39, 0.49] \\
SimLingo           & 63.0          & 0.28 [0.19, 0.36] \\
BridgeDrive        & 71.8          & 0.26 [0.18, 0.35] \\
\midrule
\multicolumn{3}{@{}l}{\emph{Warehouse (Isaac Sim)}} \\
X-Mobility         & --            & 0.34 [0.30, 0.38] \\
\bottomrule
\end{tabular}

\caption{\textbf{Driving score and decision accuracy per agent.} Both
averaged with equal weight over the certified families that contain
variants where the correct call is to stop. Driving: the Charger gauntlet
(four certified blocked widths) and the two oncoming-flow sweeps (wait
windows labeled by the gap certificate; the defective shortest spacing
variant is excluded). Warehouse: the slab pincer, floor stock and composed
stacks, which have no driving score. Avg.\ DS: mean CARLA driving score.
Avg.\ DA: decision accuracy, the balanced mean of the clean-pass rate on
passable or usable variants and the stop rate on blocked or wait variants,
with a \pn{bootstrap.level_percent}{95}\% bootstrap interval over runs, of
zero width where every run on each side has the same outcome. Rows are ordered
by Avg.\ DA within each block; bold marks the best learned driving agent in
each column.}
\label{tab:summary}
\end{table}

\begin{figure*}[t]
\centering
\includegraphics{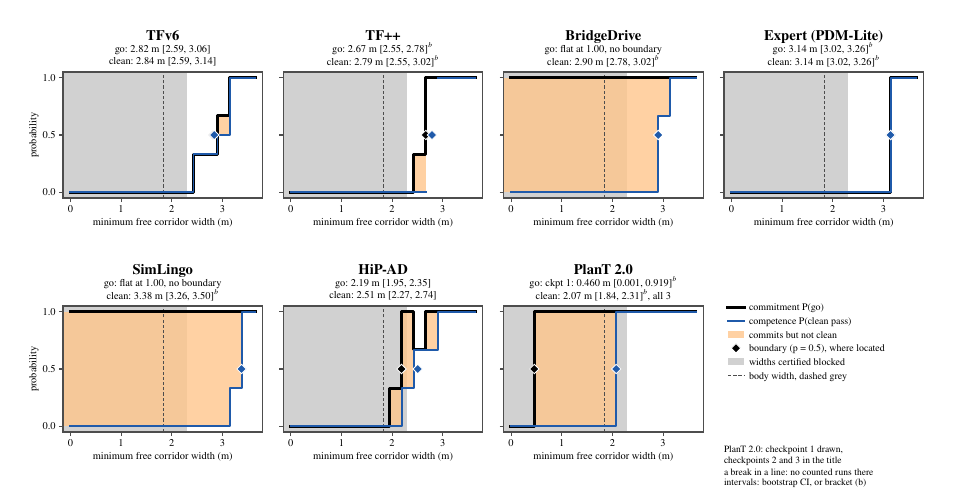}
\caption{\textbf{Failure-boundary profiles on the Charger gauntlet.} Twelve
widths, three traffic seeds per agent. Each panel is one agent. Black:
commitment, the share of runs in which the agent attempts the passage. Blue:
competence, the share of runs that pass without contact. Orange: widths where
the agent commits but does not pass cleanly. Grey: widths certified blocked,
below the \pn{fm_004_threshold_sweep.policy_floor_m}{2.30}\,m policy floor;
the dashed line marks the \pn{fm_004_threshold_sweep.body_limit_m}{1.84}\,m
body width. A diamond marks each located boundary at one half. Panel titles
give each boundary with its \pn{bootstrap.level_percent}{95}\% bootstrap 
interval or, where the outcomes are separated or the fit does not converge, 
its bracket ($^{b}$): the pair of adjacent widths between which the observed 
rate crosses one half. A title reads ``flat'' where every run has
the same outcome, ``not located in range'' where the fitted curve crosses
one half outside the tested range, and ``below tested range'' where the
observed rate stays above one half at every tested value. PlanT~2.0: checkpoint
1 is drawn; the panel title gives the boundaries of checkpoints 2 and 3 where
located.}
\label{fig:profiles-gauntlet}
\end{figure*}

We first compare the two summary scores across agents and test how
decision accuracy depends on the certification constants. We then report
each family with its per-agent profiles, the CARLA families first and the
warehouse last.

\paragraph{Overview.}
\label{sec:results-overview}

\autoref{tab:summary} reports the driving score and decision accuracy,
each averaged over the certified families that contain variants where the
correct call is to stop. Among the driving agents the expert has the
highest decision accuracy (\pn{pdm_lite_expert.decision_accuracy}{0.74},
\pn{bootstrap.level_percent}{95}\% interval
\pn{pdm_lite_expert.decision_accuracy}{[0.67, 0.81]}) and TFv6 the highest of
the learned agents (\pn{lead_regnety032.decision_accuracy}{0.60}
\pn{lead_regnety032.decision_accuracy}{[0.53, 0.68]}). The two intervals
overlap, as do those of TFv6 and PlanT~2.0 at its first checkpoint
(\pn{plant2.decision_accuracy}{0.56}
\pn{plant2.decision_accuracy}{[0.50, 0.61]}), so the order at the top should
be read as a point estimate. The driving score ranks the first three rows of \autoref{tab:summary} in
the same order as decision accuracy, but not the rest. BridgeDrive has the fourth-highest
driving score (\pn{bridgedrive.avg_driving_score}{71.8}) but the lowest
decision accuracy of any driving agent
(\pn{bridgedrive.decision_accuracy}{0.26}). It never stops at a blocked width,
and passes cleanly in only
\pn{bridgedrive.clean_passes_on_passable_and_usable}{21 of its 40} runs on
passable widths and usable windows. TF++, which stops at every blocked width
of the gauntlet, has the lowest driving score
(\pn{tfpp.avg_driving_score}{59.5}) and a decision accuracy of
\pn{tfpp.decision_accuracy}{0.49}. On the gauntlet's blocked widths TF++ and
TFv6 stop on all \pn{fm_004_threshold_sweep.tfpp.blocked.runs,
fm_004_threshold_sweep.lead_regnety032.blocked.runs}{12} and HiP-AD on
\pn{fm_004_threshold_sweep.hipad.blocked.correct_stop}{11 of 12}, while
BridgeDrive and SimLingo never stop on one, and PlanT~2.0 stops on
\pn{fm_004_threshold_sweep.plant2.blocked.correct_stop}{3 of 12} at one checkpoint and on
none at the other two.
 On the oncoming-flow wait windows few learned agents
wait: TFv6 waits on \pn{lead_regnety032.waits_on_wait_windows}{6 of its 29}
windows, BridgeDrive on \pn{bridgedrive.waits_on_wait_windows}{3} and every
other learned agent on none, against
\pn{pdm_lite_expert.waits_on_wait_windows}{13 of 29} for the expert. In the
warehouse, X-Mobility scores \pn{x_mobility.decision_accuracy}{0.34}. It
passes cleanly in
\pn{x_mobility.clean_passes_on_passable_and_usable}{40 of its 115} runs on
passable gaps and stops at blocked gaps only on the composed stacks. A score
of \pn{decision_accuracy.one_sided_score}{0.50} marks an agent that passes
cleanly on every passable variant and never stops, as PlanT~2.0 does at its
third checkpoint; a score below \pn{decision_accuracy.one_sided_score}{0.50}
is lower than an agent that stops everywhere would receive. The driving score
cannot separate stopping from continuing, because continuing past a blockage
with contact scores as well as or better than stopping in front of it
(\autoref{tab:case-study-split}).

\textbf{Robustness to the certification constants.} Both labels rest on
declared constants: the gauntlet's
\pn{fm_004_threshold_sweep.policy_floor_m}{2.30}\,m policy floor and the gap
certificate's \pn{flow_families.certificate.margin_s}{0.5}\,s margin.
Re-labeling the flow windows with margins from
\pn{sensitivity.margin_s.values}{0 to 1}\,s, and separately counting the
\pn{fm_004_threshold_sweep.blocked_axis_range}{2.07}\,m gauntlet width, which
is wider than the car but narrower than the policy floor, as passable, leaves
both ends of the ranking unchanged. The expert stays first
(\pn{pdm_lite_expert.decision_accuracy_band}{0.70 to 0.83}), TFv6 second
(\pn{lead_regnety032.decision_accuracy_band}{0.58 to 0.64}), and SimLingo and
BridgeDrive last (\pn{bridgedrive.decision_accuracy_band}{0.22 to 0.33}); only
the middle group, PlanT~2.0's checkpoints, TF++ and HiP-AD
(\pn{sensitivity.middle_group}{0.43 to 0.56}), changes order. As at the
\pn{flow_families.certificate.margin_s}{0.5}\,s margin, each traffic seed
keeps its own window, so at every margin a seed is labeled from the gap it
actually offered, and the seeds of one variant may fall into different bands.

The failure-boundary profiles show where each agent's commitment exceeds its
competence. No single agent is best on every family: SimLingo passes the
narrowest corridor of any agent in the corridor-width family yet needs the
widest corridor of any agent on the Charger gauntlet, while HiP-AD, the
last to pass cleanly in the corridor-width family, passes the gauntlet
more narrowly than any other sensor-based agent
(\pn{fm_004_threshold_sweep.hipad.p_clean.boundary}{2.51}\,m). A profile
therefore describes an agent on a particular family, and does not reduce to
one aggregate score. The following paragraphs report each family in turn.

\begin{figure*}[t]
\centering
\includegraphics{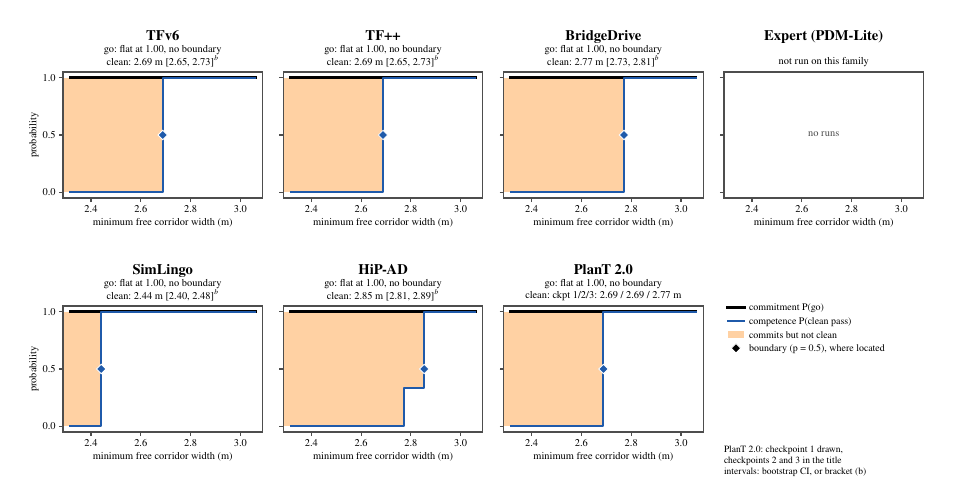}
\caption{\textbf{Failure-boundary profiles on the corridor-width family.} Ten widths, three traffic seeds; every variant is certified passable. Panels and colors as in \autoref{fig:profiles-gauntlet}. The expert does not register these barriers and was not run.}
\label{fig:profiles-pincer}
\end{figure*}

\begin{figure*}[t]
\centering
\includegraphics{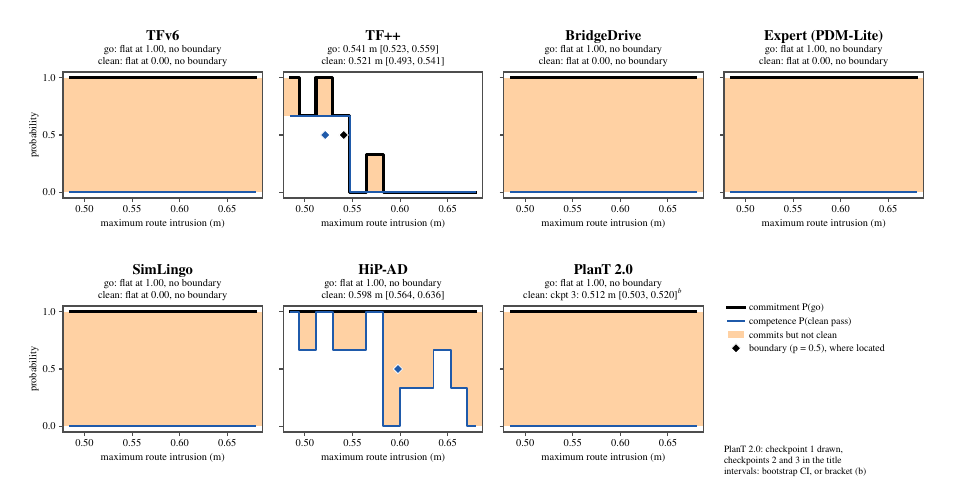}
\caption{\textbf{Failure-boundary profiles on the machine intrusion.} Twelve
intrusion depths, three traffic seeds; every variant is certified passable,
the deepest leaving \pn{fm_001_intrusion_sweep.min_free_corridor_m}{2.32}\,m
of free corridor. Each panel is one agent. Black: commitment, the share of
runs in which the agent attempts the passage. Blue: competence, the share of
runs that pass without contact. Orange: depths where the agent commits but
does not pass cleanly. Panel titles give each boundary with its interval.}
\label{fig:profiles-machine}
\end{figure*}

\paragraph{Charger gauntlet.}
\label{sec:results-gauntlet}

Two parked cars narrow the corridor from
\pn{fm_004_threshold_sweep.axis_range}{3.64}\,m to fully closed; the four
narrowest widths are certified blocked (\autoref{fig:profiles-gauntlet}). The
expert commits only where it can pass: both of its boundaries sit at
\pn{fm_004_threshold_sweep.pdm_lite_expert.p_go.boundary}{3.14}\,m (bracket
\pn{fm_004_threshold_sweep.pdm_lite_expert.p_go.boundary}{[3.02, 3.26]}),
identical at all three seeds. TFv6 behaves the same way at a narrower width,
with commitment and competence boundaries
\pn{fm_004_threshold_sweep.lead_regnety032.boundary_gap}{2\,cm} apart
(\pn{fm_004_threshold_sweep.lead_regnety032.p_go.boundary}{2.82}\,m
\pn{fm_004_threshold_sweep.lead_regnety032.p_go.boundary}{[2.59, 3.06]} and
\pn{fm_004_threshold_sweep.lead_regnety032.p_clean.boundary}{2.84}\,m
\pn{fm_004_threshold_sweep.lead_regnety032.p_clean.boundary}{[2.59, 3.14]});
it stops in all \pn{fm_004_threshold_sweep.lead_regnety032.blocked.runs}{12}
of its blocked runs and refuses
\pn{fm_004_threshold_sweep.lead_regnety032.passable.refusal}{8 of its 24} runs
on passable widths. TF++ commits earlier than it can execute (commitment
\pn{fm_004_threshold_sweep.tfpp.p_go.boundary}{2.67}\,m, competence
\pn{fm_004_threshold_sweep.tfpp.p_clean.boundary}{2.79}\,m).

HiP-AD also has a commitment boundary
(\pn{fm_004_threshold_sweep.hipad.p_go.boundary}{2.19}\,m
\pn{fm_004_threshold_sweep.hipad.p_go.boundary}{[1.95, 2.35]}) and stops in
\pn{fm_004_threshold_sweep.hipad.blocked.correct_stop}{11 of its 12} blocked
runs. BridgeDrive and SimLingo commit at every width, including the closed
corridor, and pass cleanly only from
\pn{fm_004_threshold_sweep.bridgedrive.p_clean.boundary}{2.90}\,m and
\pn{fm_004_threshold_sweep.simlingo.p_clean.boundary}{3.38}\,m respectively;
below that they continue with contact, and BridgeDrive displaces a parked car
by up to
\pn{fm_004_threshold_sweep.bridgedrive.max_obstacle_displacement_m}{55}\,m.
The driving score can reward driving through a blocked corridor: on the
blocked widths SimLingo has the highest mean score of any agent
(\pn{fm_004_threshold_sweep.simlingo.score_split}{52.2}), while the agents
that stop in all their blocked runs score between
\pn{fm_004_threshold_sweep.pdm_lite_expert.score_split}{37.8} and
\pn{fm_004_threshold_sweep.tfpp.score_split}{40.6}. BridgeDrive, which also
never stops, is the exception, with a mean of
\pn{fm_004_threshold_sweep.bridgedrive.score_split}{36.1} on the same widths,
below every agent that stops (\autoref{tab:case-study-split}).

\begin{table}[t]
\centering
\footnotesize
\setlength{\tabcolsep}{3pt}
\begin{tabular}{@{}lcccccc@{}}
\toprule
Object & Expert & TFv6 & TF++ & BridgeD. & SimLingo & HiP-AD \\
\midrule
Vending machine  & 3/3   & \cellcolor{orange!25}1/3\,$\cdot$\,2R & 3/3   & \cellcolor{orange!25}1/3\,$\cdot$\,2L & 3/3   & 3/3 \\
Plastic chair    & 3/3   & \cellcolor{orange!25}1/3\,$\cdot$\,2R & 3/3   & 3/3                                   & 3/3   & 3/3 \\
\midrule
12 other objects & 36/36 & 36/36                                 & 36/36 & 36/36                                 & 36/36 & 36/36 \\
\bottomrule
\end{tabular}

\caption{\textbf{Identity ladder.} Fourteen roadside objects at one
certified intrusion (\pn{fm_003_asset_swap.max_route_intrusion_m}{0.5}\,m,
about \pn{fm_003_asset_swap.min_free_corridor_m}{8}\,m free corridor), three
traffic seeds each. Cells give clean passes out of runs; R: refusal, L: late
commit; shading marks cells with fewer clean passes than runs. The twelve
other objects are the dog house, street barrier, traffic warning, mailbox,
playground slide, plastic table, traffic cone, trampoline, barbecue grill,
iron plank, waste bin and cardboard box.}
\label{tab:identity}
\end{table}
\begin{figure*}[t]
\centering
\includegraphics{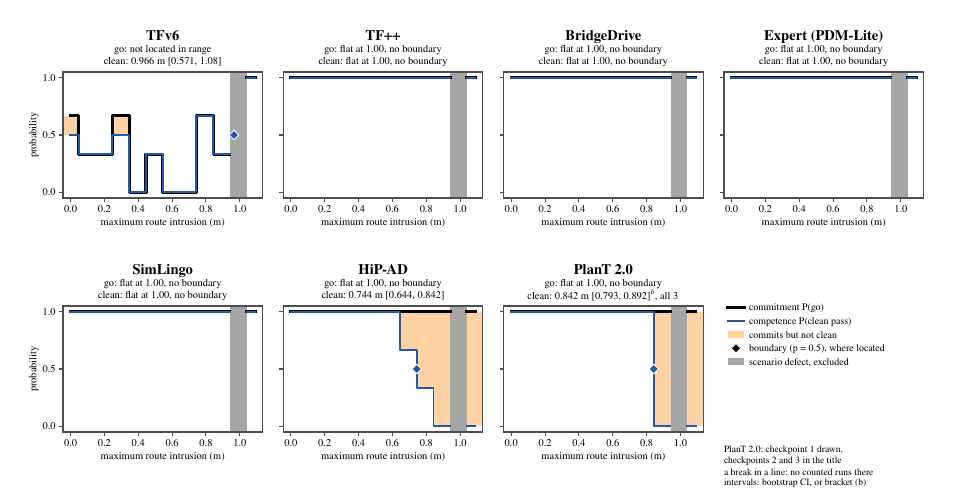}
\caption{\textbf{Failure-boundary profiles on the cluttered intrusion.} Twelve
intrusion depths, three traffic seeds; every variant is certified passable.
Panels and colors as in \autoref{fig:profiles-gauntlet}; intrusion grows to
the right. Grey: the \pn{fm_005_hero.defect_variant_x}{0.99}\,m variant, a
scenario defect (the vending machine tips over the moment it is placed, before
any car arrives), excluded from every score.}
\label{fig:profiles-clutter}
\end{figure*}

\paragraph{Corridor width.}
\label{sec:results-pincer}

Two street barriers, moved in from the right against a vending machine on
the left curb, narrow the corridor from
\pn{fm_002_corridor_pincer.axis_range}{3.06}\,m to
\pn{fm_002_corridor_pincer.axis_range}{2.32}\,m; every width is certified
passable (\autoref{fig:profiles-pincer}). Every agent commits at every width,
so the family separates agents by competence alone. SimLingo passes cleanly
down to \pn{fm_002_corridor_pincer.simlingo.p_clean.boundary}{2.44}\,m, the
narrowest of any agent; TFv6, TF++ and two of PlanT~2.0's checkpoints from
\pn{fm_002_corridor_pincer.lead_regnety032.p_clean.boundary,
fm_002_corridor_pincer.tfpp.p_clean.boundary,
fm_002_corridor_pincer.plant2.p_clean.boundary,
fm_002_corridor_pincer.plant2_s2.p_clean.boundary}{2.69}\,m; BridgeDrive and
PlanT's third checkpoint from
\pn{fm_002_corridor_pincer.bridgedrive.p_clean.boundary,
fm_002_corridor_pincer.plant2_s3.p_clean.boundary}{2.77}\,m; and HiP-AD from
\pn{fm_002_corridor_pincer.hipad.p_clean.boundary}{2.85}\,m. Below these
widths agents scrape the barriers rather than stop. Some clean passes are also
near misses:
\pn{fm_002_corridor_pincer.near_misses.sensor_roster}{15 of the 42} clean
passes of TFv6, TF++ and BridgeDrive cleared a barrier by less than the mirror
allowance (\pn{fm_002_corridor_pincer.mirror_allowance_m}{0.13}\,m) yet scored
\pn{fm_002_corridor_pincer.near_misses.sensor_roster}{86 to 97}. The expert
does not register these barriers and is not run.

\paragraph{Machine intrusion.}
\label{sec:results-machine}

A vending machine intrudes \pn{fm_001_intrusion_sweep.axis_range}{0.49} to
\pn{fm_001_intrusion_sweep.axis_range}{0.68}\,m into the lane, always leaving
at least \pn{fm_001_intrusion_sweep.min_free_corridor_m}{2.32}\,m of free
corridor (\autoref{fig:profiles-machine}). The expert, whose demonstrations
trained TFv6 and BridgeDrive, contacts the machine in all
\pn{fm_001_intrusion_sweep.pdm_lite_expert.runs}{36} of its runs; TFv6,
BridgeDrive and SimLingo do the same, and PlanT~2.0 passes cleanly in only
\pn{fm_001_intrusion_sweep.plant2.checkpoints.passable.clean_thread}{6 of 108}
runs. Two agents behave differently. TF++ passes cleanly at the shallowest
depths and refuses the deeper ones
(\pn{fm_001_intrusion_sweep.tfpp.passable.clean_thread}{8} clean passes,
\pn{fm_001_intrusion_sweep.tfpp.passable.refusal}{25} refusals,
\pn{fm_001_intrusion_sweep.tfpp.passable.contact_thread}{3} contacts), with
commitment and competence boundaries at
\pn{fm_001_intrusion_sweep.tfpp.p_go.boundary}{0.54} and
\pn{fm_001_intrusion_sweep.tfpp.p_clean.boundary}{0.52}\,m. HiP-AD passes
cleanly in \pn{fm_001_intrusion_sweep.hipad.passable.clean_thread}{20 of 36}
runs (competence boundary
\pn{fm_001_intrusion_sweep.hipad.p_clean.boundary}{0.60}\,m
\pn{fm_001_intrusion_sweep.hipad.p_clean.boundary}{[0.56, 0.64]}). Among TFv6,
TF++, BridgeDrive and the expert, the driving score ranks the expert, which
contacts the machine in every run, highest
(\pn{fm_001_intrusion_sweep.pdm_lite_expert.mean_driving_score}{63.5}), and
TF++, the only one of the four that ever passes it cleanly, lowest
(\pn{fm_001_intrusion_sweep.tfpp.mean_driving_score}{50.8}), largely because
TF++ refuses the deeper variants.

\begin{figure*}[t]
\centering
\includegraphics{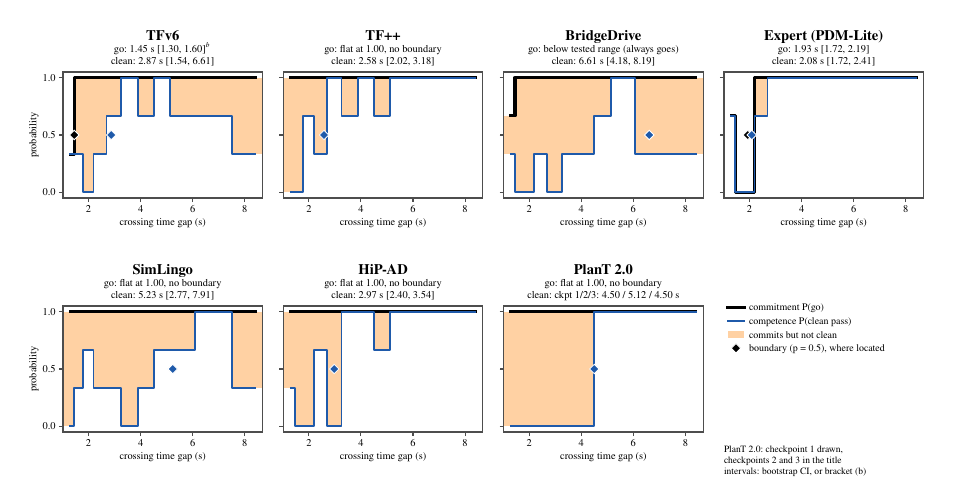}
\caption{\textbf{Failure-boundary profiles on the oncoming-flow speed sweep.} Eleven crossing time gaps, three traffic seeds; larger gaps are easier. Panels and colors as in \autoref{fig:profiles-gauntlet}. A go is entry into the oncoming lane while the flow is active.}
\label{fig:profiles-speed}
\end{figure*}

\begin{figure*}[t]
\centering
\includegraphics{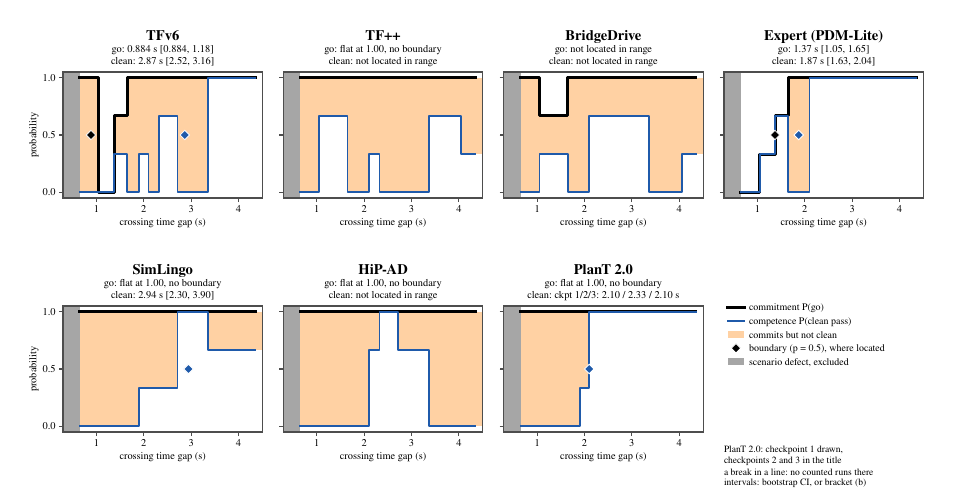}
\caption{\textbf{Failure-boundary profiles on the oncoming-flow spacing
sweep.} Eleven crossing time gaps, three traffic seeds; larger gaps are
easier. Panels and colors as in \autoref{fig:profiles-gauntlet}. A go is entry
into the oncoming lane while the flow is active. Grey: the shortest spacing
variant, at \pn{fm_012_source_dist_interval_sweep.defect_variant_x}{0.45}\,s, a
scenario defect excluded from every score.}
\label{fig:profiles-spacing}
\end{figure*}

\paragraph{Cluttered intrusion.}
\label{sec:results-clutter}

A vending machine and two waste bins intrude up to
\pn{fm_005_hero.axis_range}{1.09}\,m into an otherwise open road
(\autoref{fig:profiles-clutter}). TF++, SimLingo and the expert pass cleanly
in every run, and BridgeDrive in
\pn{fm_005_hero.bridgedrive.passable.clean_thread}{34 of 36}, with two late
commitments. TFv6 refuses in
\pn{fm_005_hero.lead_regnety032.passable.refusal}{20 of its 36} runs, starting
at the shallowest variants where nothing intrudes into the lane, which
suggests that it reacts to the cluttered scene rather than to the intrusion
itself. It passes cleanly in every run only at the deepest variant. The
second-deepest variant (\pn{fm_005_hero.defect_variant_x}{0.99}\,m) is a
scenario defect: the vending machine tips over the moment the harness places
it, in every run and before any car arrives, so it is excluded from every
score. HiP-AD's and PlanT~2.0's competence boundaries on the remaining
variants are \pn{fm_005_hero.hipad.p_clean.boundary}{0.74}\,m
\pn{fm_005_hero.hipad.p_clean.boundary}{[0.64, 0.84]} and
\pn{fm_005_hero.plant2.p_clean.boundary,
fm_005_hero.plant2_s2.p_clean.boundary,
fm_005_hero.plant2_s3.p_clean.boundary}{0.84}\,m
\pn{fm_005_hero.plant2.p_clean.boundary,
fm_005_hero.plant2_s2.p_clean.boundary,
fm_005_hero.plant2_s3.p_clean.boundary}{[0.79, 0.89]}.

\begin{figure*}[t]
\centering
\includegraphics{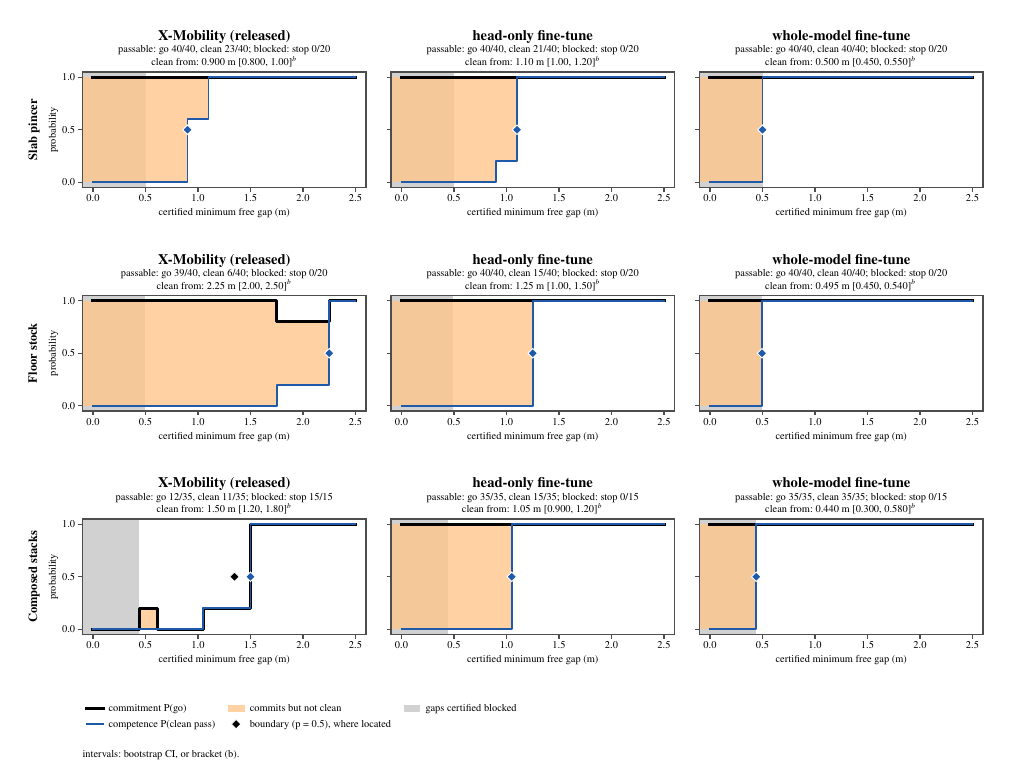}
\caption{\textbf{Failure-boundary profiles in the warehouse.} Three warehouse
families in Isaac Sim, each narrowing a robot's aisle to a single gap from
\pn{warehouse.axis_range}{2.5}\,m to a closed aisle, with the neighboring
aisle open as a detour; twelve certified gaps on the slab pincer and the floor
stock, ten on the composed stacks, five policy samples each. Rows: families.
Columns: the released X-Mobility checkpoint and its two fine-tunes, both
trained on demonstrations from the slab pincer and the floor stock; the
composed stacks are held out. The released checkpoint and the fine-tunes were
measured on the same gaps of each family. Black: commitment, the share
of samples that attempt the gap. Blue: competence, the share that pass through
it without contact. Orange: gaps the robot attempts but does not pass cleanly.
Grey: gaps certified blocked. A diamond marks each located boundary at one
half. Panel titles
give the go and clean-pass counts on passable gaps, the stop count on blocked
gaps, and the competence boundary with its interval or bracket ($^{b}$).}
\label{fig:profiles-warehouse}
\end{figure*}

\paragraph{Identity ladder.}
\label{sec:results-identity}

Fourteen roadside objects are placed at one certified intrusion,
\pn{fm_003_asset_swap.max_route_intrusion_m}{0.5}\,m into the lane with about
\pn{fm_003_asset_swap.min_free_corridor_m}{8}\,m of free corridor, so the
intrusion and the free corridor are the same for every object
(\autoref{tab:identity}); the objects themselves differ in size and shape.
TFv6 refuses the vending machine and the plastic chair at two of three seeds
each, while passing the other twelve objects cleanly at every seed. With three
seeds per object, these are
\pn{fm_003_asset_swap.lead_regnety032.passable.refusal}{4} refusals in
\pn{fm_003_asset_swap.lead_regnety032.runs}{42} runs. This suggests that the
object rather than the intrusion changes TFv6's decision, but more samples per
object would be needed to confirm it. BridgeDrive late-commits on the vending
machine at two of three seeds. The expert, TF++, SimLingo and HiP-AD pass all
fourteen objects in every run, so the effect depends on the agent and not only
on the objects.

\paragraph{Oncoming flow.}
\label{sec:results-flow}

Two families sweep the time gap in a single-lane oncoming flow at a
signalized left turn across oncoming traffic, once by flow speed and 
once by car spacing
(Figs.~\ref{fig:profiles-speed} and~\ref{fig:profiles-spacing}). The
oncoming flow is the most challenging setting in the evaluation, both for
the agents and for locating a boundary: agents typically cannot time the
crossing consistently, and their profiles are far from monotone. Most
agents commit at nearly every gap, so their profiles differ in competence.
On the speed sweep the expert commits only where it can pass (commitment
\pn{fm_011_flow_speed_sweep.pdm_lite_expert.p_go.boundary}{1.93}\,s,
competence
\pn{fm_011_flow_speed_sweep.pdm_lite_expert.p_clean.boundary}{2.08}\,s);
competence boundaries are located for TF++
(\pn{fm_011_flow_speed_sweep.tfpp.p_clean.boundary}{2.58}\,s), HiP-AD
(\pn{fm_011_flow_speed_sweep.hipad.p_clean.boundary}{2.97}\,s) and PlanT~2.0
(\pn{fm_011_flow_speed_sweep.plant2.p_clean.boundary,
fm_011_flow_speed_sweep.plant2_s3.p_clean.boundary}{4.50} to
\pn{fm_011_flow_speed_sweep.plant2_s2.p_clean.boundary}{5.12}\,s across
checkpoints), while TFv6, BridgeDrive and SimLingo collide at easy gaps as
well as hard ones, so their competence boundaries carry intervals that span
most of the tested range
(\pn{fm_011_flow_speed_sweep.lead_regnety032.p_clean.boundary}{2.87}\,s
\pn{fm_011_flow_speed_sweep.lead_regnety032.p_clean.boundary}{[1.54, 6.61]},
\pn{fm_011_flow_speed_sweep.bridgedrive.p_clean.boundary}{6.61}\,s
\pn{fm_011_flow_speed_sweep.bridgedrive.p_clean.boundary}{[4.18, 8.19]} and
\pn{fm_011_flow_speed_sweep.simlingo.p_clean.boundary}{5.23}\,s
\pn{fm_011_flow_speed_sweep.simlingo.p_clean.boundary}{[2.77, 7.91]}). On the
spacing sweep the expert's competence boundary is
\pn{fm_012_source_dist_interval_sweep.pdm_lite_expert.p_clean.boundary}{1.87}\,s,
TFv6's
\pn{fm_012_source_dist_interval_sweep.lead_regnety032.p_clean.boundary}{2.87}\,s
and SimLingo's
\pn{fm_012_source_dist_interval_sweep.simlingo.p_clean.boundary}{2.94}\,s.
Against the gap certificate, the flow families show the same pattern as the
gauntlet, with a moving obstacle. Every agent goes on every usable window, but
on the \pn{flow_families.wait_windows_per_agent}{29} wait windows per agent
the expert waits \pn{pdm_lite_expert.waits_on_wait_windows}{13} times, TFv6
\pn{lead_regnety032.waits_on_wait_windows}{6} times, BridgeDrive
\pn{bridgedrive.waits_on_wait_windows}{3} times and every other learned agent
never. The expert also crosses cleanly on every usable window, while most
learned agents fail to cross cleanly on some of them; HiP-AD crosses cleanly
on none of its six usable spacing windows. Decision accuracy on the flow
families is therefore
\pn{fm_011_flow_speed_sweep.pdm_lite_expert.decision_accuracy}{0.79} and
\pn{fm_012_source_dist_interval_sweep.pdm_lite_expert.decision_accuracy}{0.68}
for the expert and
\pn{flow_families.learned.decision_accuracy_range}{0.00 to 0.62} for the
learned agents. The conditional windows, where only a rolling crossing fits,
separate the agents the other way: PlanT~2.0 and HiP-AD take every one with a
rolling entry and SimLingo all but one, while the expert takes none, because
it always enters from rest. The shortest spacing variant is a scenario defect,
in which the oncoming cars collide with each other before the ego arrives, and
is excluded from every score.

\begin{figure*}[t]
    \centering
    \includegraphics{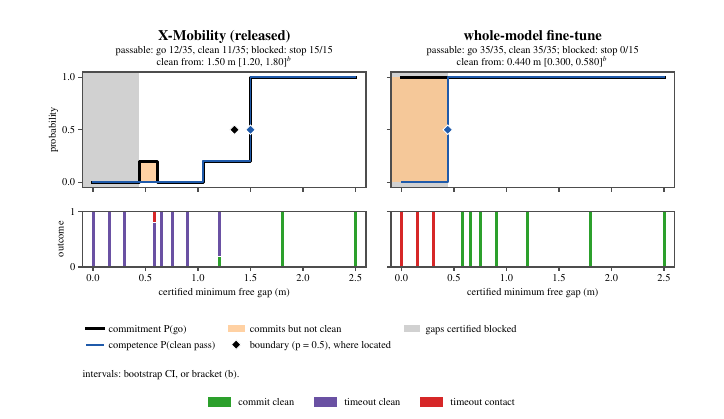}
    \caption{\textbf{X-Mobility before and after fine-tuning, on the held-out
    composed stacks.} Five policy samples per certified gap for the released
    checkpoint (left) and the whole-model fine-tune (right), on the
    same ten gaps. Top: commitment (black), the share of samples that decide
    to go, and competence (blue), the share that pass cleanly, against the
    certified minimum free gap; diamonds mark the located
    boundaries, orange the gaps where the robot commits but does not pass
    cleanly, and grey the gaps certified blocked. Panel titles give
    the go and clean-pass counts on passable gaps, the stop count on blocked
    gaps, and the competence boundary with its bracket. Bottom: the outcome of
    each sample at each gap: commit clean, a passage without contact; timeout
    clean, the robot neither passes nor touches anything before the time limit,
    its way of stopping; timeout contact, the time limit reached after a
    contact.}
    \label{fig:xmobility-finetune}
\end{figure*}

\paragraph{Warehouse.}
\label{sec:results-warehouse}

The three warehouse families narrow a robot's aisle to a single gap, from
\pn{warehouse.axis_range}{2.5}\,m down to a closed aisle, while the
neighboring aisle stays open (\autoref{fig:profiles-warehouse}, left column).
The scripted expert takes every passable gap and detours at every blocked one,
\pn{fm_isaac_001.expert.correct_calls,fm_isaac_003.expert.correct_calls}{12 of 12}
correct calls on the slab pincer and the floor stock and
\pn{fm_isaac_004.expert.correct_calls}{10 of 10} on the composed stacks. The
released X-Mobility checkpoint fails differently in each family. On the slab
pincer it commits at every gap and passes cleanly only from
\pn{fm_isaac_001.x_mobility.p_clean.boundary}{0.90}\,m (bracket
\pn{fm_isaac_001.x_mobility.p_clean.boundary}{[0.80, 1.00]});
\pn{fm_isaac_001.x_mobility.contacts}{37 of its 60} samples in the family end
in contact. On
the floor stock it commits at
\pn{fm_isaac_003.x_mobility.go_on_passable}{39 of 40} passable samples but
passes cleanly only from
\pn{fm_isaac_003.x_mobility.p_clean.boundary}{2.25}\,m (bracket
\pn{fm_isaac_003.x_mobility.p_clean.boundary}{[2.00, 2.50]}). On the composed
stacks it commits only from
\pn{fm_isaac_004.x_mobility.commitment_per_gap}{1.8}\,m and touches a stack
once in \pn{fm_isaac_004.x_mobility.runs}{50} samples: it waits in front of
nearly every narrower gap, so it stops at all
\pn{fm_isaac_004.x_mobility.blocked.runs}{15} blocked samples but passes
cleanly on only \pn{fm_isaac_004.x_mobility.pass_side}{11 of 35} passable
ones. The robot never takes the open neighboring aisle: across all three
families, released or fine-tuned, it never attempts a detour, and either
commits or waits. The two curves and decision accuracy apply to this
simulator, robot and policy without any changes.

\subsection{Re-measuring an Agent Across Training}
\label{sec:results-training}

Because a family is defined by a measured axis rather than a fixed route,
the same profiles can be re-measured on each new version of an agent. We
illustrate this with two fine-tunes of the released X-Mobility checkpoint
(\autoref{fig:profiles-warehouse}, middle and right columns). A \emph{head-only} fine-tune adapts the policy heads, and a
\emph{whole-model} fine-tune adapts the full network. Both are trained on
demonstrations from the slab pincer and the floor stock; the composed
stacks are held out.

Both fine-tunes pass more passable gaps cleanly in total. The head-only 
fine-tune lowers the competence boundary by about a meter on the floor stock
(from \pn{fm_isaac_003.x_mobility.p_clean.boundary}{2.25} to
\pn{fm_isaac_003.x_mobility_fusion_v3.p_clean.boundary}{1.25}\,m) and by half
a meter on the held-out composed stacks (from
\pn{fm_isaac_004.x_mobility.p_clean.boundary}{1.50} to
\pn{fm_isaac_004.x_mobility_fusion_v3.p_clean.boundary}{1.05}\,m), but raises
it on the slab pincer, from \pn{fm_isaac_001.x_mobility.p_clean.boundary}{0.90}
to \pn{fm_isaac_001.x_mobility_fusion_v3.p_clean.boundary}{1.10}\,m, passing
\pn{fm_isaac_001.x_mobility_fusion_v3.pass_side}{21 of 40} cleanly against the
released checkpoint's \pn{fm_isaac_001.x_mobility.pass_side}{23 of 40}. 
The whole-model fine-tune passes every passable gap cleanly in
all three families, including the held-out one. On blocked gaps, however, both
get worse: neither stops at a single blocked gap in any family. The change is
largest on the held-out composed stacks (\autoref{fig:xmobility-finetune}).
There the released checkpoint waits in front of nearly every gap below
\pn{fm_isaac_004.x_mobility.commitment_per_gap}{1.8}\,m, stopping at all
\pn{fm_isaac_004.x_mobility.blocked.runs}{15} blocked samples but passing only
\pn{fm_isaac_004.x_mobility.pass_side}{11 of 35} passable ones. The
whole-model fine-tune commits at every gap, passes every passable one cleanly,
and goes on all \pn{fm_isaac_004.x_mobility_vis_full.blocked.runs}{15} blocked
samples, every one of which ends in contact
(\pn{fm_isaac_004.x_mobility_vis_full.wait_side}{0 of 15} stops). The head-only
fine-tune also goes on all
\pn{fm_isaac_004.x_mobility_fusion_v3.blocked.runs}{15}.

Decision accuracy alone would not reveal this. The whole-model fine-tune's
mean rises from \pn{x_mobility.decision_accuracy}{0.34} to
\pn{x_mobility_vis_full.decision_accuracy}{0.50}, the value of an agent that
passes every passable gap cleanly and never stops. On the held-out stacks it
falls (\pn{fm_isaac_004.x_mobility.decision_accuracy}{0.66} to
\pn{fm_isaac_004.x_mobility_vis_full.decision_accuracy}{0.50}), because a
perfect passable side replaces a perfect blocked side. The head-only
fine-tune's mean falls to \pn{x_mobility_fusion_v3.decision_accuracy}{0.22}.
The gain in clean passage and the loss of the decision to stop only become
visible when the two sides are read separately. The released checkpoint and
the fine-tunes are measured on the same certified gaps of each family
(twelve, twelve and ten), so the boundaries compare directly. The blocked-side
change, from \pn{fm_isaac_004.x_mobility.wait_side}{15 of 15} stops to none,
holds at every blocked gap.
\paragraph{Code and data.} The measurement and certification stage of the
pipeline, the ten scenario families and the records of the seventeen
campaigns are in the public repository, at tag 
\href{https://github.com/speedym/failure-boundary-profiles/tree/v1.0-paper}{v1.0-paper}. 
The simulator output of every campaign and the scored results computed from 
it are on the Hugging Face Hub (DOI  \href{https://doi.org/10.57967/hf/11195}{10.57967/hf/11195}), one archive per 
campaign, named by the hash of its contents and pinned by the records. Every 
measured figure, table and number in this paper is computed from these files 
by the released code, and the scenario family figure is cut from the recordings. 
The warehouse runner, the CARLA map and the agents' checkpoints are not part of
the release; the driving agents are available from their authors.
\section{Conclusion}

The driving score used by closed-loop CARLA benchmarks does not separate
an agent's decision to stop from its decision to continue. On the Charger
gauntlet, an agent that never stops at a blocked width scores higher than
the agents that always stop, while a second agent that never stops scores
lower than them. Failure-boundary profiles address this by sweeping a
single certified factor and estimating commitment and competence along it.
This separates the decision to attempt a maneuver from the ability to carry
it out, locates where each changes, and reports that location with an
interval that accounts for the simulator's run-to-run variance. Decision
accuracy then credits a clean pass where the certified label says go and a
stop where it says stop.

Compared with the expert, the learned driving agents differ most on the
stop side. They wait for at most
\pn{lead_regnety032.waits_on_wait_windows}{6 of 29} gaps in oncoming traffic,
against \pn{pdm_lite_expert.waits_on_wait_windows}{13} for the expert. On
passable variants, several of them pass cleanly more often than the expert,
which is conservative on the gauntlet and hits the vending machine in every
run. The warehouse policy stops at blocked gaps in only one of three families.
Since the families are defined by measured axes rather than fixed routes, the
same profiles can be measured again for each new version of an agent, to see
whether a boundary moved, in which direction and by how much. For two
fine-tunes of the warehouse policy, this shows better clean passage together
with the loss of every stop (\autoref{sec:results-training}). We think this is
where the method is most useful: telling developers where their agent's
commitment exceeds its competence, and whether their usual metric would have
caught it.

\paragraph{Limitations.}
All of our results come from simulation. They compare agents within CARLA
and within Isaac Sim, and do not yet tell us whether the same failure
boundaries appear on real vehicles or robots.

The labels that decision accuracy relies on are conventions we state
explicitly rather than facts we can measure without assumptions. Whether a
corridor counts as blocked depends on the policy floor, and whether a gap
in oncoming traffic counts as usable depends on the margin of the gap
certificate. Within the range we tested, changing these constants leaves
the strongest and weakest agents in place but reorders the agents in
between, so the middle of the ranking should be read with care. The
summary itself rests on only three families that contain a stop side, and
the expert we compare against is a scripted planner that serves as a
reference, not as an upper bound on good driving.

The precision of the profiles is limited by the number of runs behind
them. With three seeds per variant, the intervals of neighboring agents
often overlap, and boundaries that lie closer together than their
intervals should not be read as different. More runs near each boundary
would help, but evaluation is already costly: running all seven driving
agents takes roughly \pn{driving_agents.wall_hours}{220 hours} on a single
GPU, which may be out of reach for teams with limited compute. The warehouse
study is narrower still. It covers one robot and one policy in three versions,
and the two fine-tuned versions were trained on two of its three families, so
only the composed stacks are a held-out test.

{
    \small
    \bibliographystyle{ieeenat_fullname}
    \bibliography{main}
}

\end{document}